\documentclass{article}

 \usepackage[nonatbib, final]{neurips_2020_ml4ad}

\usepackage[utf8]{inputenc} 
\usepackage[T1]{fontenc}    
\usepackage{hyperref}       
\usepackage{url}            
\usepackage{booktabs}       
\usepackage{amsfonts}       
\usepackage{nicefrac}       
\usepackage{microtype}      
\usepackage{graphicx}
\usepackage{multirow}
\graphicspath{{images/}}

\title{PePScenes: A Novel Dataset and Baseline for Pedestrian Action Prediction in 3D}

\author{%
  Amir Rasouli \And Tiffany Yau \And Peter Lakner \And Saber Malekmohammadi \And Mohsen Rohani \And Jun Luo \And \\
 Noah's Ark Laboratory\\
 Huawei, Canada\\
  \texttt{\{amir.rasouli,tiffany.yau1, peter.lakner,saber.malekmohammadi,}\\
    \texttt{mohsen.rohani,jun.luo1\}@huawei.com} \\
}

\begin{document}

\maketitle

\begin{abstract}
Predicting the behavior of road users, particularly pedestrians, is vital for safe motion planning in the context of autonomous driving systems.  Traditionally, pedestrian behavior prediction has been realized in terms of forecasting future trajectories. However, recent evidence suggests that predicting higher-level actions, such as crossing the road, can help improve trajectory forecasting and planning tasks accordingly. There are a number of existing datasets that cater to the development of pedestrian action prediction algorithms, however, they lack certain characteristics, such as bird's eye view semantic map information, 3D locations of objects in the scene, etc., which are crucial in the autonomous driving context. To this end, we propose a new pedestrian action prediction dataset created by adding per-frame 2D/3D bounding box and behavioral annotations to the popular autonomous driving dataset, nuScenes. In addition, we propose a hybrid neural network architecture that incorporates various data modalities for predicting pedestrian crossing action. By evaluating our model on the newly proposed dataset, the contribution of different data modalities to the prediction task is revealed. The dataset is available at \url{https://github.com/huawei-noah/PePScenes}.
\end{abstract}

\vspace{-0.2cm}
\section{Introduction}
\vspace{-0.2cm}

One of the major challenges faced by autonomous driving systems is predicting road users' behavior, in particular, pedestrians as they exhibit a diverse set of actions \cite{rasouli_2017_IV} influenced by various environmental and social factors \cite{rasouli_2019_ITS}. In the context of driving, behavior prediction is commonly actualized in terms of forecasting the future trajectories of road users. However, as the recent developments in this field suggest, prediction of higher-level actions of road users, e.g. pedestrian crossing actions, can be beneficial for trajectory forecasting and motion planning  \cite{Chaabane_2020_WACV,Rasouli_2019_BMVC,Rasouli_2019_ICCV,Liang_2019_CVPR,Casas_2018_CORL}.

In recent years, a number of pedestrian action prediction algorithms have been introduced \cite{rasouli_2020_arxiv} many of which were trained and evaluated on existing pedestrian behavior datasets \cite{Malla_2020_CVPR,Liu_2020_RAL,Rasouli_2019_ICCV,Rasouli_2017_ICCVW}. These datasets, however, are limited since they do not contain information such as 3D maps of environments, 3D locations of objects, etc. necessary for prediction in the context of autonomous driving systems. 

In this paper, we introduce a novel dataset for pedestrian crossing action and dense trajectory prediction for autonomous driving applications.  Our dataset contains new per-frame bounding box and behavioral annotations for the nuScenes dataset \cite{Caesar_2020_CVPR}. The annotations are added to 3D as well as 2D data making it suitable for various applications in the autonomous driving domain.

 Furthermore, we propose a hybrid baseline model that uses multi-modal data inputs to predict pedestrian crossing action. We train and evaluate the proposed model on our new dataset and show how different modalities of data contribute to prediction accuracy.
\vspace{-0.2cm}
\section{Related Works}
\vspace{-0.2cm}
\subsection{Datasets}
\vspace{-0.2cm}

Pedestrian behavior prediction can take two forms: implicit where pedestrian trajectories are forecasted and explicit where pedestrian actions are predicted. There are many existing datasets that cater to trajectory prediction for different domains such as surveillance \cite{Liang_2020_CVPR_2,Robicquet_2016_ECCV,Lerner_2007_CGF}, anomaly detection \cite{Lu_2013_ICCV,Liu_2018_CVPR,Loy_2009_BMVC}, and intelligent driving \cite{Chang_2019_CVPR, Sun_2020_CVPR_3,Geiger_2012_CVPR}. However, the choices for pedestrian action prediction are more limited. There are a few datasets that provide rich behavioral tags along with temporally coherent spatial annotations that can be used for pedestrian action prediction in the driving context. One of the early datasets is Joint Attention in Autonomous Driving (JAAD) \cite{Rasouli_2017_ICCVW} which consists of 346 video clips annotated with 2D bounding boxes for pedestrians and behavioral tags for a subset of them along with the ego-vehicle driver's actions. A major drawback of this dataset is the lack of ego-motion information which is vital for prediction from a moving camera perspective. A more recent dataset, Pedestrian Intention Estimation (PIE) \cite{Rasouli_2019_ICCV}, rectifies this issue by providing the ego-vehicle motion parameters in addition to more samples, annotations for all relevant objects (besides pedestrians), and pedestrian intention information obtained by conducting a human experiment. There are two other datasets similar to PIE, namely Trajectory Inference using Targeted Action priors Network (TITAN) \cite{Malla_2020_CVPR} and Stanford-TRI Intent Prediction (STIP) \cite{Liu_2020_RAL} both of which provide 2D bounding box and pedestrian behavior annotations. These datasets, however, are available under very restrictive terms of use.  VIENA$^2$ \cite{Aliakbarian_2018_ACCV} is another action anticipation dataset which contains only simulated video sequences collected from a computer game. 

The major drawback of the existing pedestrian action prediction datasets is the lack of information, such as the semantic map of the environment, 3D coordinates, etc. all of which are necessary for developing algorithms for autonomous driving systems. Given that our proposed dataset is built on an existing autonomous driving dataset, all required types of data are available.
\vspace{-0.2cm}
\subsection{Behavior Prediction in Driving}
\vspace{-0.2cm}

The dominant approach to predicting road users' behaviors is to forecast their future trajectories \cite{Fang_2020_CVPR,Liang_2020_CVPR,Phan-Minh_2020_CVPR,Chandra_2019_CVPR,Rhinehart_2018_ECCV}. However, recent evidence suggests that predicting high-level actions of road users can benefit various planning tasks both directly \cite{Chaabane_2020_WACV,Rasouli_2019_BMVC,Gujjar_2019_ICRA} and indirectly, e.g. via improving trajectory forecasting \cite{Malla_2020_CVPR,Rasouli_2019_ICCV,Liang_2019_CVPR,Casas_2018_CORL}.

Many algorithms have been proposed for pedestrian action prediction. A subset of these algorithms relies on feedforward architectures \cite{Chaabane_2020_WACV,Saleh_2019_ICRA,Gujjar_2019_ICRA,Rasouli_2017_ICCVW}.  Some of these models predict actions directly by classifying various components in the scene \cite{Saleh_2019_ICRA}, some predict from intermediate features such as pedestrian head orientation \cite{Rasouli_2017_ICCVW}, and others generate future scene representations which are used to classify future actions \cite{Gujjar_2019_ICRA}. Opposed to such unimodal approaches are recurrent architectures that benefit from a combination of different data modalities, such as images, ego-motion information, poses, trajectories, etc. to make predictions \cite{Mangalam_2020_WACV,Liu_2020_RAL,Rasouli_2019_BMVC,Aliakbarian_2018_ACCV}.

While feedforward networks are very powerful for capturing the spatiotemporal representations of the scenes, recurrent networks provide flexibility for combining multi-modal data with different dimensionalities. In our proposed approach, we take advantage of both of these architectures in a hybrid framework that uses both convolutional layers for processing image data and recurrent networks for encoding trajectories and ego-motion information.

\vspace{-0.2cm}
\section{PePScenes Dataset}
\vspace{-0.2cm}

The proposed dataset is a set of additional 2D/3D bounding box and behavioral annotations to the existing nuScenes dataset \cite{Caesar_2020_CVPR}. Although the main goal of creating this dataset was for pedestrian action prediction, the newly added annotations can be used in various tasks such as tracking, trajectory prediction, object detection, etc. We refer to the new data as \textbf{Pe}destrian \textbf{P}rediction on nu\textbf{Scenes} (PePScenes). 

\textbf{Annotations.} nuScenes has 1000 segments (i.e. data sequences) out of which annotations for 850 are available online. We added bounding box annotations for all the existing objects in the annotated portion of the dataset. However, for behavioral annotations, we only chose a subset of samples that, 1) appear in front of the ego-vehicle, 2) have or appear to have an intention of crossing (e.g. they are not far away on a sidewalk), and 3) are observable for at least a few frames prior to making crossing decision. Given these criteria, we added behavioral labels to 719 unique pedestrian tracks. The overall statistics of the proposed dataset can be found in Table \ref{data_stats}. 

\textbf{Bounding boxes.} nuScenes contains LIDAR scans and camera images recorded at $20$ and $12$Hz respectively. The existing bounding box annotations of nuScenes, however, are at $2$Hz which is fairly sparse, especially for the task of pedestrian behavior prediction. As a result, we augmented spatial annotations of pedestrians and all objects at 10Hz. We interpolated the bounding boxes between two consecutive original annotations using the global coordinates of pedestrians in the environment. To better align the new bounding boxes with the actual samples, we used a 2D detection algorithm, RetinaNet \cite{Lin_2017_CVPR} pre-trained on COCO \cite{Lin_2014_ECCV}, to first localize pedestrians in the images and then use the detected boxes to adjust the locations of added bounding boxes according to the projection of 3D bounding boxes on the image plane.  In the end, we randomly sub-sampled a portion of the data and manually evaluated them to assure the quality of newly added boxes. 

\textbf{Behavioral labels.} Behavioral labels for crossing actions were added to a subset of pedestrians. Each of the unique pedestrian samples has an object-level annotation indicating whether at a given point in the sequence they will cross the road in front of the ego-vehicle. In addition, for each frame, we also include the current crossing state of the pedestrian, i.e. whether they are currently crossing or not by specifying the start and end time of crossing events. For samples that eventually cross the road, a label is added to specify the critical point in time when crossing starts.

\vspace{-0.2cm}
\section{Proposed Model}
\vspace{-0.2cm}

\textbf{Problem statement.} We formulate pedestrian action prediction as an optimization process in which the goal is to learn distribution $p(A_i^{t+m}| SC_{o}, M_{o},  L_{o}, V_o)$ for some pedestrian $ 1 < i < n$ where $A_i^{t+m} \in \{0,1\}$ is pedestrian crossing action at some time $t+m$ in the future. Predictions are based on observed scenes  $SC_o = \{sc^{1}, sc^{2}, ..., sc^{t}\}$, changes in the semantic map of the environment $M_o = \{m^{1}, m^{2}, ..., m^{t}\}$, the pedestrian's observed trajectory $L_o =\{l^{1}, l^{2}, ..., l^{t}\}$ and the ego-vehicle states $ V = \{v^{1}, v^{2}, ..., v^{t}\}$. 

\begin{figure*}
\centering{
\includegraphics[width=0.85\textwidth]{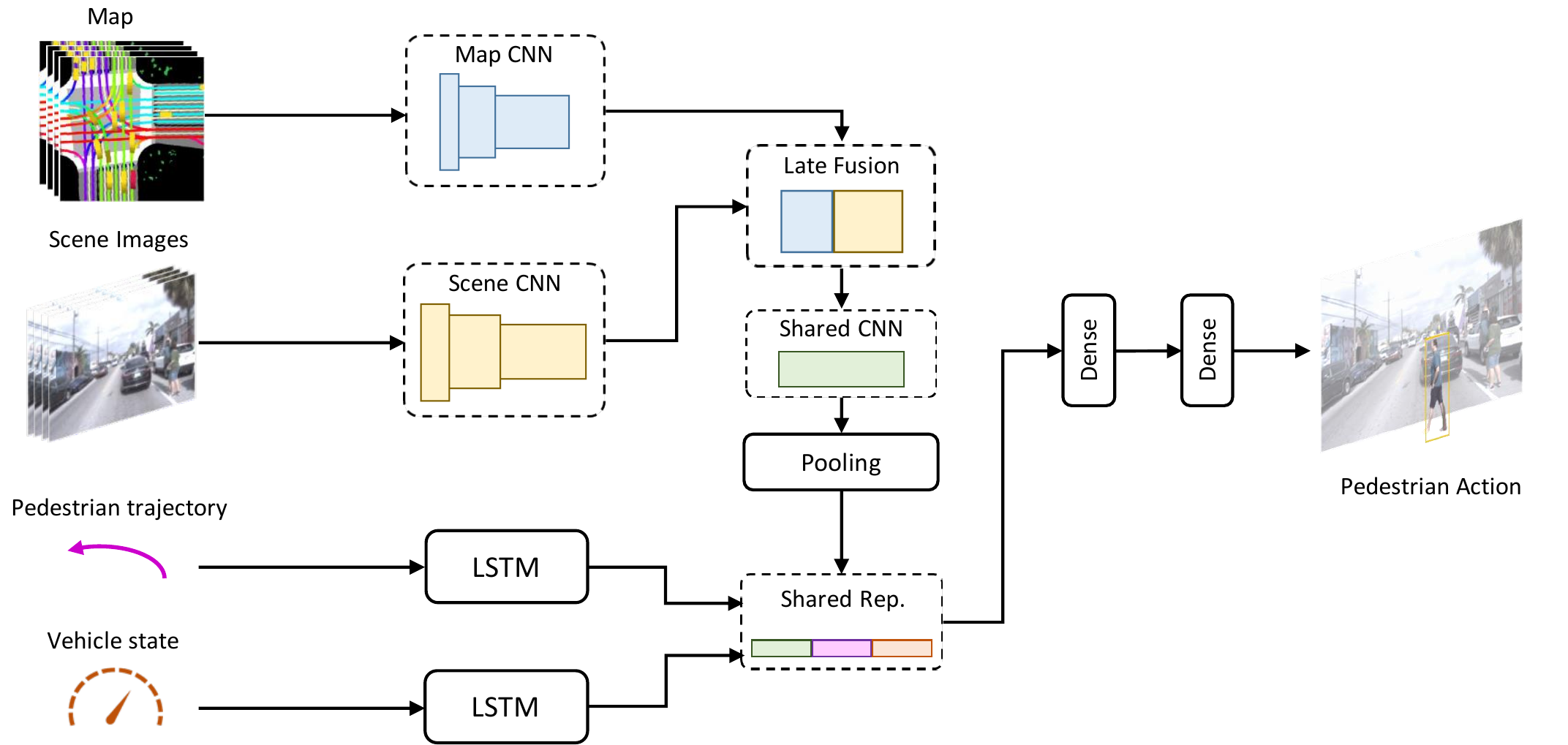}
\caption{An overview of the proposed architecture. The model relies on four different input modalities: semantic maps, scene images, trajectories and ego-vehicle states. Visual features are processed with two sets of Conv2D layers followed by a late-fusion Conv2D layer for joint processing. Trajectories and ego-vehicle states are processed using two LSTMs the output of which are concatenated to visual features to form a shared representation which is fed into consecutive dense layers to make predictions.}
\label{main_diagram}}
\vspace{-0.5cm}
\end{figure*}

\textbf{Architecture}. As mentioned earlier, we employ a hybrid approach to encode different input modalities (see Figure \ref{main_diagram}). We use rasterized maps encoded as 3-channel images similar to \cite{Cui_2019_ICRA}. The map is of size $30 \times 30$ meters centered around the ego-vehicle. As for scene images, we use the entire scene images from the forward center camera resized to $300 \times 300$ pixels. Both map and scene image sequences are stacked  channel-wise and fed into two separate sets of Conv2D layers with sizes $\{[32, 3, 3], [64, 3, 2], [128, 3, 2]\}, \{[64, 3, 3], [128, 3, 2], [256, 3, 2]\}$ respectively where values in order stand for \textit{[number of filters, kernel size, stride]}. The final outputs of map and scene conv layers are concatenated and fed into a single Conv2D layer, $[512,3,1]$, followed by a global average pooling to generate visual representations.

For trajectories, we use $[x,z]$ coordinates of pedestrians in the environment and ego-vehicle state represented by velocity $[v_x, v_y, v_z]$. Both trajectories and ego-vehicle states are processed using two LSTMs with $128$ cells. The final shared representation is formed by concatenating the output of the LSTMs and visual representations. The shared representation is then fed into two dense layers, with dropout of $0.5$ in between, to predict actions. For learning, we use binary cross-entropy loss function.

\vspace{-0.2cm}
\section{Evaluation}
\vspace{-0.2cm}
\textbf{Data.} We split the data into train/test sets with a ratio of $70/30$ while maintaining the ratios of positive and negative samples consistent. Following \cite{Rasouli_2019_BMVC}, we clip sequences up to the first frame of crossing events. In cases where no crossing occurs, we select the last frame in the center-view camera where the pedestrian is visible. We choose an observation length of $0.5$ seconds (or $5$ frames at $10$Hz) and sample sequences from each pedestrian track between $1$ to $2s$ to the event of crossing with an overlap of $50\%$ between each sample. 

\textbf{Training.} We trained the model end-to-end using RMSProp \cite{Tieleman_2012_tech} optimizer with batch size of $8$ and learning rate of $5\times 10^{-5}$ for $50$ epochs. To compensate for data imbalance, we used class weights based on the ratio of positive and negative samples.

\textbf{Metrics.} For evaluation purposes common binary classification metrics as in \cite{Rasouli_2019_BMVC} are used including $accuracy$, Area Under the Curve ($AUC$), $F1$, and $precision$. 
\vspace{-0.2cm}
\subsection{Crossing Prediction}
\vspace{-0.2cm}
We compare the performance of the proposed model to a baseline LSTM model trained only on trajectories and state-of-the-art crossing prediction algorithm, SF-GRU \cite{Rasouli_2019_BMVC}. For a fair comparison, we use global coordinates and velocity instead of 2D bounding box coordinates and the ego-vehicle speed originally used in SF-GRU. In addition, to highlight the contributions of different data modalities to the prediction task, different subsets of the proposed model are evaluated on PePScenes. We refer to these subsets based on the types of input modalities that are used.

As shown in Table \ref{results}, when relying merely on dynamics features such as trajectory the model performs poorly. By combining visual features with dynamics information, the results on all metrics show improvements. The best results are achieved on all metrics when all sources of information are included as shown by the performance of the proposed model.

\begin{table}[]
\centering{
\begin{minipage}{0.5\linewidth}
\caption{The overall statistics of the annotations. The numbers under \textit{New} column refer to the newly added annotations and under \textit{Original}, the existing nuScenes annotations.}
\resizebox{1\textwidth}{!}{
\begin{tabular}{l|cc|c}
\multicolumn{1}{c|}{Annt.} & \textit{New}                      & \multicolumn{1}{l|}{\textit{Original}} & Total \\ \hline
\# Ped. with beh.          & 719                      & -                             & 719   \\
\# Cross. peds             & 149                      & -                             & 149   \\
\# Non-cross peds.         & 570                      & -                             & 570   \\
\# Per-frame beh. annt.    & 63.4K                    & -                             & 63.4K \\
\# Ped. box annt.          & 845K                     & 222K                          & 1.06M \\
\# Other box annt          & 3.58M                    & 944K                          & 4.52M \\
Annt. frame rate           & \multicolumn{1}{l}{10Hz} & 2Hz                           & 10Hz 
\end{tabular}
}
\label{data_stats}
\end{minipage}
}
\hfill
\centering{
\begin{minipage}{0.47\linewidth}
\caption{The performance of the proposed model trained and tested on the new PePScenes dataset. Our model is evaluated with different input modalities.}
\resizebox{1\textwidth}{!}{
\begin{tabular}{ll|cccc}
  \multicolumn{2}{l|}{Method}         & Acc & AUC & F1 & Prec \\ \hline
  \multicolumn{2}{l|}{LSTM}       & 0.78  & 0.54  & 0.20   &     0.39       \\
  \multicolumn{2}{l|}{SF-GRU \cite{Rasouli_2019_BMVC}}   & 0.86  & 0.60   & 0.31   & 0.39       \\\hline
\multirow{5}{*}{Ours} & Scene     & 0.80    &  0.58  & 0.26   & 0.24         \\
					  & Map     &   0.82  & 0.55    &  0.26  &   0.23         \\
					  & Map+Scene  &  0.85  & 0.62   & 0.35   &   0.38          \\
					  & Map+Scene+Traj &  0.86  & 0.62    &  0.34  &  0.43         \\ 
					  & \textbf{All} &   \textbf{0.87}   & \textbf{0.71}    &   \textbf{0.48} &   \textbf{0.47}     
\end{tabular}
}
\label{results} 
\end{minipage}

}
\vspace{-0.5cm}
\end{table}
\vspace{-0.2cm}

\section{Conclusion}
\vspace{-0.2cm}

We proposed a novel dataset for pedestrian behavior prediction by augmenting the nuScenes dataset with more than 60K behavioral and 4 million bounding box annotations. This is the first dataset that provides high-level action annotations on 3D data for research in pedestrian behavior prediction. In addition, new dense annotations in the dataset are suitable for tasks such as tracking, detection, trajectory prediction, etc. We also proposed a hybrid model for pedestrian crossing prediction and showed how a combination of different data modalities can improve the accuracy of prediction.

\bibliographystyle{IEEEtran}
\bibliography{refs}

\end{document}